\def\year{2020}\relax
\documentclass[letterpaper]{article} 
\usepackage{hyperref}
\usepackage{aaai20}  
\usepackage{times}  
\usepackage{helvet} 
\usepackage{courier}  
\usepackage{graphicx} 
\usepackage{subfigure}
\usepackage{graphicx}  
\usepackage{todonotes}
\usepackage{xcolor}
\usepackage{amssymb}
\usepackage{amsmath}
\usepackage{bbm}
\usepackage{array}
\usepackage{natbib}
\usepackage[ruled, lined, linesnumbered, commentsnumbered, longend]{algorithm2e}
\usepackage{booktabs}
\usepackage{multirow}
\DeclareUnicodeCharacter{00D7}{*}

\title{Mixup Without Hesitation}
\author{
Hao Yu\textsuperscript{\rm 1}, Huanyu Wang\textsuperscript{\rm 1}, Jianxin Wu\textsuperscript{\rm 1} \\ \textsuperscript{\rm 1}State Key Laboratory for Novel Software Technology, Nanjing University \\ \{yuh, wanghy\}@lamda.nju.edu.cn, wujx2001@gmail.com}

\begin{document}

\maketitle

\begin{abstract}
Mixup linearly interpolates pairs of examples to form new samples, which is easy to implement and has been shown to be effective in image classification tasks. However, there are two drawbacks in mixup: one is that more training epochs are needed to obtain a well-trained model; the other is that mixup requires tuning a hyper-parameter to gain appropriate capacity but that is a difficult task. In this paper, we find that mixup constantly explores the representation space, and inspired by the exploration-exploitation dilemma in reinforcement learning, we propose mixup Without hesitation (mWh), a concise, effective, and easy-to-use training algorithm. We show that mWh strikes a good balance between exploration and exploitation by gradually replacing mixup with basic data augmentation. It can achieve a strong baseline with less training time than original mixup and without searching for optimal hyper-parameter, i.e., mWh acts as mixup without hesitation. mWh can also transfer to CutMix, and gain consistent improvement on other machine learning and computer vision tasks such as object detection. Our code is open-source and available at \href{https://github.com/yuhao318/mwh}{https://github.com/yuhao318/mwh} 
\end{abstract}

\section{Introduction}
Deep learning has made great breakthroughs in various computer vision problems such as image classification (\citealt{deng2009imagenet,simonyan2014very,krizhevsky2012imagenet}) and object detection (\citealt{ren2015faster}). However, requiring lots of training data is the well-known drawback of deep learning, and data augmentation methods have partially alleviated this difficulty.

In particular, mixup, proposed by Zhang et al. (\citeyear{zhang2017mixup}), is based on virtual examples created by linearly interpolating two random samples and their corresponding labels, i.e.,
\begin{equation}
  \label{eq1}
  \begin{aligned}
    \tilde{x}&=\lambda x_{i}+(1-\lambda) x_{j} \\
    \tilde{y}&=\lambda y_{i}+(1-\lambda) y_{j}
  \end{aligned}
\end{equation}
where $x_i$ and $x_j$ are two data samples, and $y_i$ and $y_j$ are their labels. The mixing coefficient $\lambda$ is sampled from a beta distribution $\mathbf{Beta}(\alpha, \alpha)$. Mixup allows deep learning models to train on a large number of virtual examples resulting from the random combination, and the standard cross-entropy loss is calculated on the soft-labels $\tilde{y}$ instead of hard labels. Therefore, mixup involves the representation space unseen during normal training, and raises the generalization of deep neural networks significantly, especially on small datasets (\citealt{thulasidasan2019mixup}).

However, it is also well-known that mixup suffers from slow convergence and requires a sophisticated selection of the hyper-parameter $\alpha$. In detail, 
\begin{itemize}
  \item Mixup requires more epochs to converge. Since it explores more regions of the data space, longer training is required, e.g., it takes 200 epochs to train ResNet-50 on ImageNet with mixup (\citealt{zhang2017mixup}), but a normal training routine of 90 epochs is sufficient. Note that this observation not only exists in mixup, but can also be found in other data augmentation methods that strongly increase the complexity of training data (\citealt{chen2020gridmask, yun2019cutmix}).
	\item Mixup requires an $\alpha$ value to sample mixing coefficients. Different $\alpha$ values usually lead to big differences in model accuracy. Zhang et al. (\citeyear{zhang2017mixup}) mentioned that mixup improves performance better when $\alpha \in [0.1,0.4]$, and larger $\alpha$ may cause underfitting. However, $\alpha$ greater than 1 tends to perform better in some cases (\citealt{thulasidasan2019mixup}). In other words, the generalization ability of mixup is heavily affected by hyper-parameter selection, but choosing a suitable $\alpha$ is quite difficult. 
\end{itemize}

In order to solve both difficulties, we propose mWh, which stands for \textbf{m}ixup \textbf{W}ithout \textbf{h}esitation (mWh). Instead of using mixup to augment data throughout the entire model training process, mWh accelerates mixup by periodically turning the mixing operation off, which also makes it robust to the hyper-parameter $\alpha$. The contributions of mWh are:
\begin{itemize}
  \item Through carefully designed experiments and observations, we show that mixup attains its accuracy improvement through boldly \emph{exploring} the representation space, which also (unfortunately) leads to the two drawbacks. Basic data augmentation (e.g., flipping and cropping) focuses more on \emph{exploiting} the space. Hence, the proposed mWh strikes a good exploration-exploitation trade-off, and achieves both high accuracy and training speed.
  \item We gain new benchmarks of image classification consistently. Regardless of whether epochs are doubled or not, the results of mWh are better than mixup. Compared to mixup with twice longer the training time, mWh can still achieve approximately the same accuracy without large training time.
  \item mWh is robust with respect to $\alpha$. With a default $\alpha$ value, mWh performs consistently well in a variety of computer vision tasks. 
\end{itemize}

\section{Related Work}
First, we briefly review data augmentation methods and the related works that inspired this paper.

One of the important problems in computer vision is training with a small amount of data, as deep learning models often overfit with small datasets. Data augmentation is a family of techniques to solve this difficulty, and basic data augmentation methods, such as horizontal reflection, rotation and rescaling, have been widely applied to many tasks and often boost the model accuracy. Mixup can be regarded as a kind of data augmentation method and it often enhances the generalization performance of CNNs. Mixup can also ease the over-confident prediction problem for deep neural networks (\citealt{thulasidasan2019mixup}). Similar interpolation can be applied in semi-supervised learning (\citealt{berthelot2019mixmatch,li2020dividemix}), model adversarial robustness (\citealt{pang2019mixup}) and domain adaptation (\citealt{xu2020adversarial}). Based on the idea of mixup, AdaMixUp (\citealt{guo2019mixup}) proposes to learn better mixing distributions by adding a vanilla network. Manifold mixup (\citealt{verma2018manifold}) shares similarities with mixup. It trains neural networks on linear combinations of hidden representations of training examples.  

Apart from mixup, basic data augmentation can also expand the sample space, and some novel data augmentation methods have recently been proposed, too. Some data augmentation approaches are based on searching, like AutoAugment (\citealt{cubuk2019autoaugment}), Fast AutoAugment (\citealt{lim2019fast}) and PBA (\citealt{ho2019population}). AutoAugment designs a search space to combine various data augmentation strategies to form a policy, but the whole search algorithm is very computationally demanding. Some data augmentation techniques crop the input image. Cutout (\citealt{devries2017improved}) randomly masks out square regions of an input image during training. GridMask improves the existing information dropping algorithms (\citealt{chen2020gridmask}). Similar to mixup, CutMix (\citealt{yun2019cutmix}) also involves two training samples: it cuts one image patch and pastes it to another training image. He et al. (\citeyear{he2019data}) analyze the distribution gap between clean and augmented data. They preserve the standard data augmentation and refine the model with 50 epochs after training DNNs with mixup.

Note that these elaborate methods change the original images significantly, and almost always elongate the training process, e.g., in Manifold mixup, Verma et al. (\citeyear{verma2018manifold}) trained PreAct ResNet-18 for 2000 epochs in CIFAR-10 and CIFAR-100, but 100 epochs of training will be enough without Manifold mixup. CutMix and GridMask also need careful hyper-parameter selection. Although they often achieve higher accuracy than basic data augmentation, more training epochs lead to significantly inflated financial and environmental costs, which is the common and significant drawback of mixup and these methods. Hence, we propose mWh to solve this dilemma. Its goal is to achieve higher accuracy even without many epochs or hyper-parameter selection.

\section{mixup Without hesitation (mWh)}
We propose mWh in this section. First, we present a brief review of our motivations, also analyze the effect of mixup training and reveal its property. Then we propose mixup Without hesitation (mWh), a simple plug-and-play training strategy. Finally, we study the role of every stage as well as the influence of hyper-parameters in mWh.
\subsection{Motivations}
Reintroducing basic data augmentation back and replacing some mixup operation is the key in mWh, and we first establish our motivations for it.

Deep learning models often overfit seriously in small datasets, and mixup provides an effective mechanism to improve the generalization ability. But because mixup generates a large amount of virtual data with higher uncertainty than usual input data (cf. Eq. \ref{eq1}), model training needs more epochs to stabilize the representation. 
Theoretically, the epochs required for model convergence can be reduced by replacing mixup data with raw data. Right now a tendency of deep learning research is to extend training time to get higher accuracy, however, it is not easy to determine whether their accuracy improvement comes from more epochs, overfitting or some benefits of the proposed algorithms. Prolonged training also consumes more resources and needs more computations, which is not economical. Therefore, providing a better trade-off between network performance and computational burden is important. 

Meanwhile, sophisticated data augmentation methods require the selection of proper hyper-parameters. In contrast, basic data augmentation methods do not have this burden. If we reintroduce basic data augmentation, the influence of hyper-parameter selection will be reduced to some extent. Hence, our motivations are: 
\begin{itemize}
  \item First we want to speed up the model convergence.
  \item Second our method needs to be robust to the hyper-parameter selection.
\end{itemize}

Using basic and classical data augmentation to replace mixup is hence a sound choice. In order to validate our motivations, we performed some experiments to analyze the effect of mixup.

\subsection{Observations}
We investigate the role mixup plays during training, and demonstrate that with \emph{the combination of mixup and basic data augmentation}, mWh has the potential to retain the accuracy improvement brought by mixup, too. 

\begin{figure}[tbp]
  \centering  
  \includegraphics[width=0.46\textwidth]{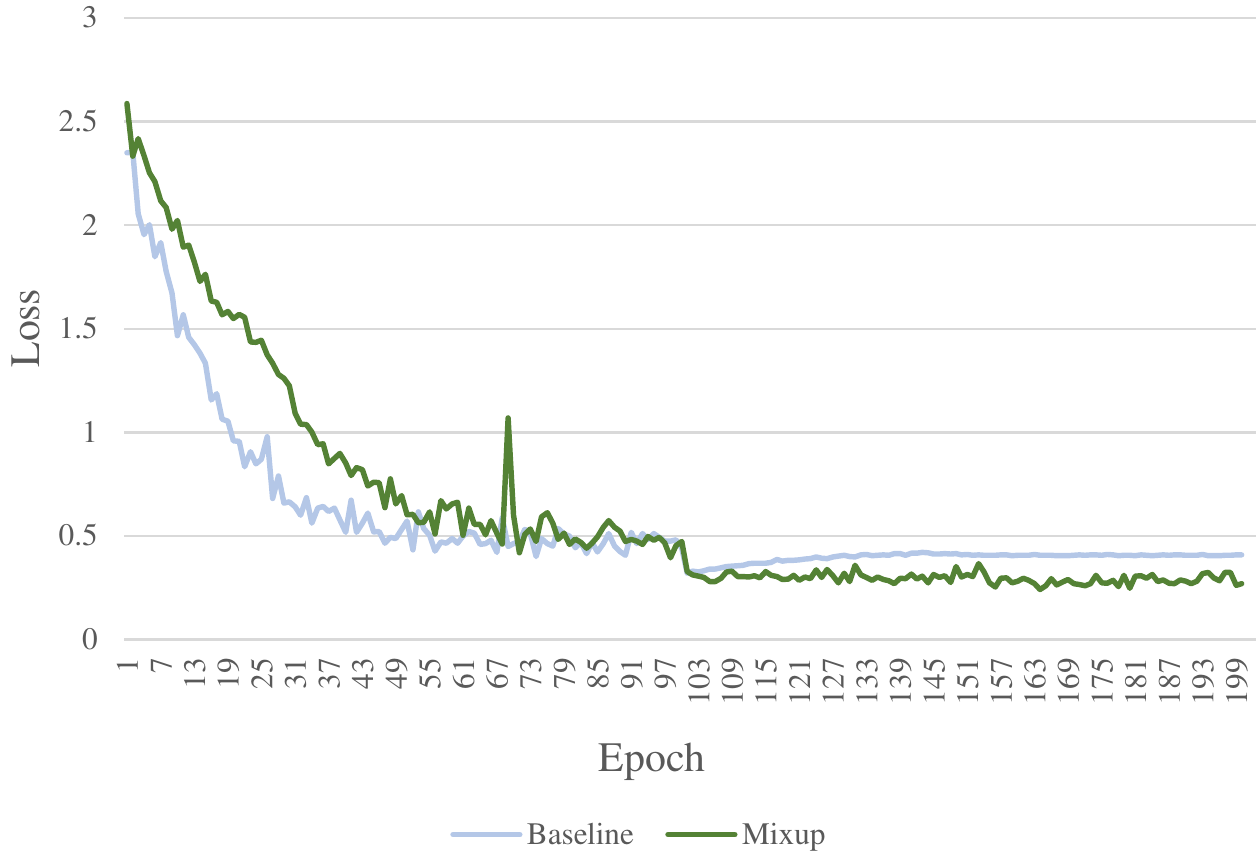}
  \caption{Cross-entropy loss on the CIFAR-10 \emph{test} set. We used PreAct ResNet-18 and $\alpha = 0.5$.}
  \label{Fig0}
\end{figure}

Training neural networks involves finding minima of a high-dimensional non-convex loss function, which can be visualized as an energy landscape. Since mixup generates more uncertain data, it enlarges the sample representation space significantly, and \emph{explores} more potential energy landscapes, so the optimization method can find a better locally optimal solution. However, the side effect of mixup is that it also brings instability during the training process. We trained PreAct ResNet-18 on CIFAR-10, and as Figure~\ref{Fig0} showed, we have two observations from the curves. The first one is: with mixup, the loss oscillates constantly on the original test dataset, but if the model is trained on the clean data, the curves are smoother. This phenomenon suggests that compared with basic data augmentation, mixup introduces higher uncertainty in the training process. The second one is: in the early training stage, mixup enlarges the test loss, which indicates it focuses on exploring the energy landscapes and will not fall into local optima prematurely. 

However, in the later training stages, since the model gets closer to convergence, exploration must not be the main goal. Instead, we should switch to \emph{exploiting} the current state using \emph{basic data augmentation}. Results in Table~\ref{table0} validate our motivations.  We train PreAct ResNet-18 on CIFAR-100 with 100 epochs and set $\alpha$ as 1.0. The result demonstrates that using mixup only in the first 50 epochs is better. Hence, we conjecture that \emph{mixup is effective because it actively explores the search space in the early epochs, while in the later epochs, mixup might be harmful}.
But, if we directly apply basic data augmentation after a model is trained with mixup (\citealt{he2019data}), this refinement operation may end up with overfitting. We trained PreAct ResNet-18 with 200 epochs in CIFAR-10 and Tiny-ImageNet-200 with mixup, and refine the model with 25 epochs without mixup. Learning rate starts at 0.1 and is divided by 10 after 100 and 150 epochs, and we set the learning rate to be the same as that in the final epochs of the last stage during refinement. The results are shown in Table~\ref{tablen}. We can observe that accuracy decreased after refinement, which indicates the number of refining epochs is difficult to control, and refining may lead to overfitting.

Putting our observations and conjectures together, we propose to \emph{gradually replace mixup with basic data augmentation} such that the learning algorithm gradually switches from exploration to exploitation, which is a good strategy to solve the \emph{exploration-exploitation dilemma} (\citealt{sutton2018reinforcement}) in our context.
\begin{table}[tbp]
  \centering
  \caption{Accuracy (\%) on CIFAR-100. Baseline means we train the model with basic data augmentation. Mixup means we apply mixup throughout the training process.  First Half Mixup means the first half of epochs apply mixup but the last do not, and similarly, Second Half Mixup means we only apply mixup in the second half of epochs. }
  \begin{tabular}{c|c|c}
  \bottomrule[1pt]
  Methods    & Top1 & Top5  \\  \hline
  Baseline    & 74.20 &	92.53 \\ 
  Mixup     & 75.25	& 92.40 \\ 
  First Half Mixup  & \textbf{75.87} &	\textbf{93.10} \\  
  Second Half Mixup  & 72.50 &	91.04 \\  
  \toprule[1pt]
  \end{tabular}
  \label{table0}
\end{table}
\begin{table}[tbp]
  \centering
  \caption{Accuracy (\%) on PreAct ResNet-18, $\alpha=0.5$. }
  \begin{tabular}{c|cc}
  \bottomrule[1pt]
  Datasets    & Mixup        &  +Refinement  \\ \hline
  CIFAR-10    & \textbf{95.46}         & 95.30  \\ 
  Tiny-ImageNet-200   & \textbf{60.73}  & 60.21  \\ 
  \toprule[1pt]
  \end{tabular}
  \label{tablen}
\end{table}
\subsection{Algorithm}
\begin{algorithm}[t]
  \caption{The mWh Training Algorithm}
  \label{alg1}
  \SetAlgoLined 
  \KwIn{Training dataset $(\mathcal{X}, \mathcal{Y}) $, number of training mini-batches $m$, two parameters $p$ and $q$ satisfying $ (0\leq p < q \leq 1)$, Beta distribution parameter $\alpha$ for mixup.}

    \For{$i=1$ to $m$}{
      Draw a mini-batch $(x_b, y_b) $.\\
      \uIf(\tcp*[f]{\textit{First stage}}){$i \leq pm$} {
          $(\tilde{x}_b,~\tilde{y}_b) = \text{mixup} ( x_b, y_b, \alpha)$ 
      }
      \uElseIf(\tcp*[f]{\textit{Second stage}}){$i \leq qm$ }{ 
        \eIf{\rm{$i$ is even} }{
          $(\tilde{x}_b,~\tilde{y}_b) = \text{mixup} (x_b, y_b, \alpha)$ 
        }{ 
          $(\tilde{x}_b,~\tilde{y}_b) = (x_b, y_b)$
        }
      }
      \Else(\tcp*[f]{\textit{Third stage}}){
        $\epsilon = \frac{ m-i}{m(1-q)}$ \\
        Randomly generate threshold $\theta \in [0,1]$.\\
        \eIf{ $\theta < \epsilon $}{
          $(\tilde{x}_b,~\tilde{y}_b)= \text{mixup}(x_b, y_b, \alpha)$ 
        }{
          $(\tilde{x}_b,~\tilde{y}_b) = (x_b, y_b)$
        }
      }
      Train model with mini-batch $(\tilde{x}_b,~\tilde{y}_b)$. 
    }
  \end{algorithm}

The mWh algorithm is in Algorithm \ref{alg1}. We use a mini-batch instead of an epoch as the unit of execution. We denote the total number of mini-batches as $m$, and use two hyper-parameters $p$ and $q~(0\leq p < q \leq 1)$ to divide the whole training process into three stages, i.e.,
\begin{itemize}
  \item First stage: from 1 to $pm$ mini-batches, we train with mixup. Note that here we assume $pm$ is an integer.
  \item Second stage: from $pm+1$ to $qm$ mini-batches, we alternate between mixup and basic data augmentation. If the last mini-batch does not apply mixup, the next one will, and vice versa.
  \item Third stage: from $qm+1$ mini-batches to the end, we run mixup with probability $\epsilon$, where $\epsilon$ decreases \emph{linearly} from 1 to 0. 
\end{itemize}

In the first stage, mWh lets the model explore a large portion of the sample representation space by consistently applying mixup.
 
The second stage is an exploration-exploitation trade-off. We periodically turn mixup on and off to avoid getting trapped in a local optimum prematurely. When we turn mixup off, the model will exploit the limited and promising region of the sample representation space with the hope of accelerating convergences. When we turn mixup on, the model will keep exploring more energy landscapes. 

In the third stage, we gradually switch to exploitation, which is inspired by the $\epsilon$-greedy algorithm (\citealt{sutton2018reinforcement}). We define an exploration rate $\epsilon$ that is initially set to 1. This rate is the probability that our model will use mixup. As $\epsilon$ decreases gradually, the model tends to choose exploitation rather than exploration.

Finding suitable values of $p$ and $q$ is essential. Note that we want mWh to be robust and insensitive to hyper-parameters. Hence, we want to fix $p$ and $q$ in \emph{all} experiments. Here we train ResNet-50 on ImageNet with 100 epochs and study the effect of different $p$ and $q$. In these experiments, the default learning rate is 0.1 with a linear warmup for the first 5 epochs and divided by 10 after training 30, 60, 90 epochs. We set batch size to 256. In Table~\ref{Figz}, we set $q$ as 0.9 and explore the impact of different $q$. We also fix $p$ as 0.6 and research the effect of $q$ in Table~\ref{Figr}. Especially, when $q$ is equal to $1.0$, we remove the third stage in mWh, and similarly, when $q$ is $0.6$, we apply the  $\epsilon$-greedy algorithm after the mini-batches of 60 percent.   Based on our experimental results, although choosing different $p$ and $q$ does not have a significant effect on the outcome, $0.6$ and $0.9$ are a reasonable choice, so we always set $p = 0.6$ and $q=0.9$. 

Now we validate our framework by an ablation study on ImageNet. We train ResNet-50 on ImageNet with 100 epochs and set $\alpha$ as 0.5. All experiments apply mixup in the top 60 percent mini-batches. Table \ref{table10} contains several results and we try different strategies in Stage 2 and Stage 3. Different rows represent using different strategies to train the model. In particular, none indicates we apply basic data augmentation. When we apply mWh at Stage 2, it refers to the alternating of mixup and basic data augmentation between the mini-batches of 60 to 90 percent. Using mWh at Stage 3 means running mixup with a probability of $\epsilon$ in the final 10 percent of the mini-batches.
\begin{table}[tbp]
  \centering
  \caption{The influence of $p$ on ImageNet. }
  \begin{tabular}{c|ccccccccccccc}
  \bottomrule[1pt]
   $p$    & $0.5$ & $0.6$ &$0.7$& $0.8$ & $0.9$ \\ \hline
  $\alpha$ = 0.2   & \textbf{76.952} & 76.948&76.814&76.856&76.894\\ 
  $\alpha$ = 0.5   & 76.782 & 76.854& \textbf{76.964}&76.754&76.736\\ \toprule[1pt]
  \end{tabular}
  \label{Figz}
\end{table} 

\begin{table}[t]
  \centering
  \caption{The influence of $q$ on ImageNet. }
  \begin{tabular}{c|ccccccccccccc}
  \bottomrule[1pt]
   $q$    & $0.6$ & $0.7$ &$0.8$& $0.9$ & $1.0$ \\ \hline
  $\alpha$ = 0.2   & 76.854 & 76.814 & 76.792 &\textbf{76.948} &76.742\\ 
  $\alpha$ = 0.5   & 76.792 & \textbf{76.942} & 76.894 &76.854 &76.716\\ \toprule[1pt]
  \end{tabular}
  \label{Figr}
\end{table}

\begin{table}[!t]
  \centering
  \caption{Accuracy (\%) of ResNet-50 trained on ImageNet.}
\begin{tabular}{cc|cc}
\bottomrule[1pt]
 Stage 2  & Stage 3 & Top1  & Top5 \\   \hline
 None & None &  76.756 &	93.314 \\
 mixup & None & 76.770  & 93.478\\
 mixup & mixup & 76.212 &	93.246 \\
 mixup & mWh & 76.832& \textbf{93.504} \\
 mWh & None & 76.772 & 	93.428 \\
 mWh & mixup & 76.388 & 93.304\\
 mWh & mWh  & \textbf{76.854} & 93.463       \\ \toprule[1pt]
\end{tabular}
\label{table10}
\end{table}

\begin{figure*}[!t]
  \centering  
  \subfigure[CIFAR-10, $\alpha=0.2$]{
  \label{Fig.sub.1}
  \includegraphics[width=0.23\textwidth]{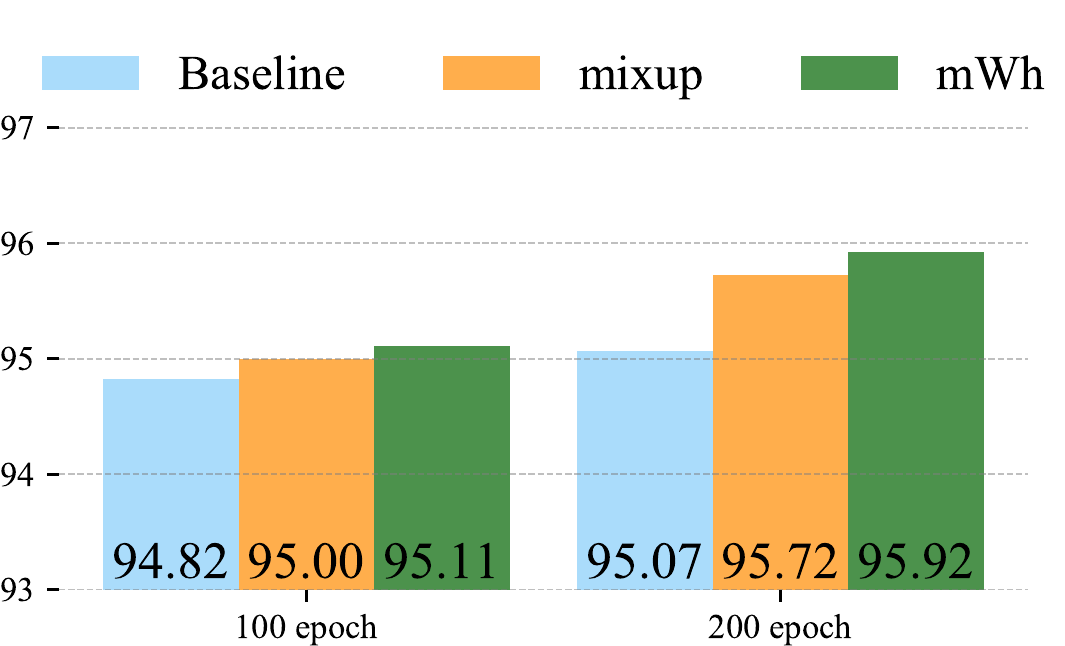}}
  \subfigure[CIFAR-10, $\alpha=0.5$]{
  \label{Fig.sub.2}
  \includegraphics[width=0.23\textwidth]{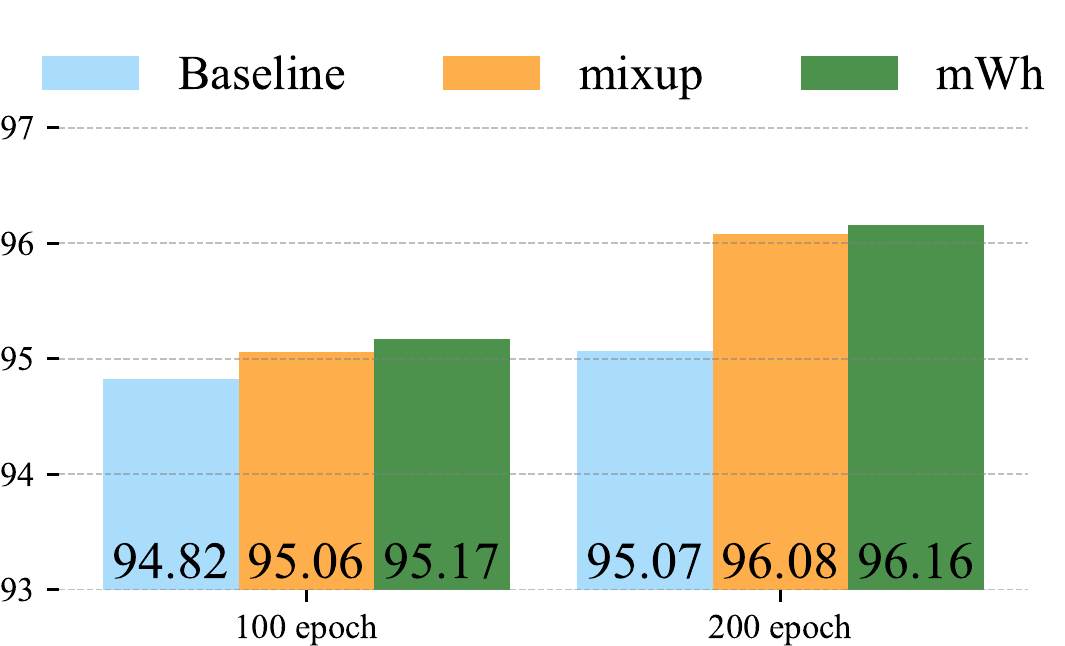}}
  \subfigure[CIFAR-100, $\alpha=0.2$]{
  \label{Fig.sub.1}
  \includegraphics[width=0.23\textwidth]{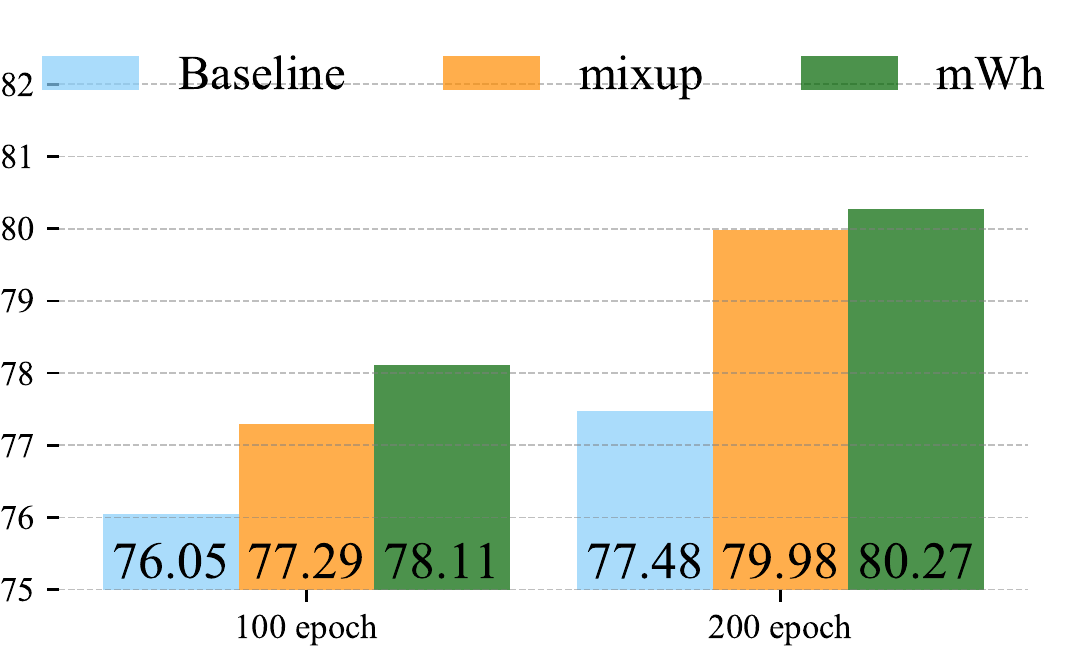}}
  \subfigure[CIFAR-100, $\alpha=0.5$]{
  \label{Fig.sub.2}
  \includegraphics[width=0.23\textwidth]{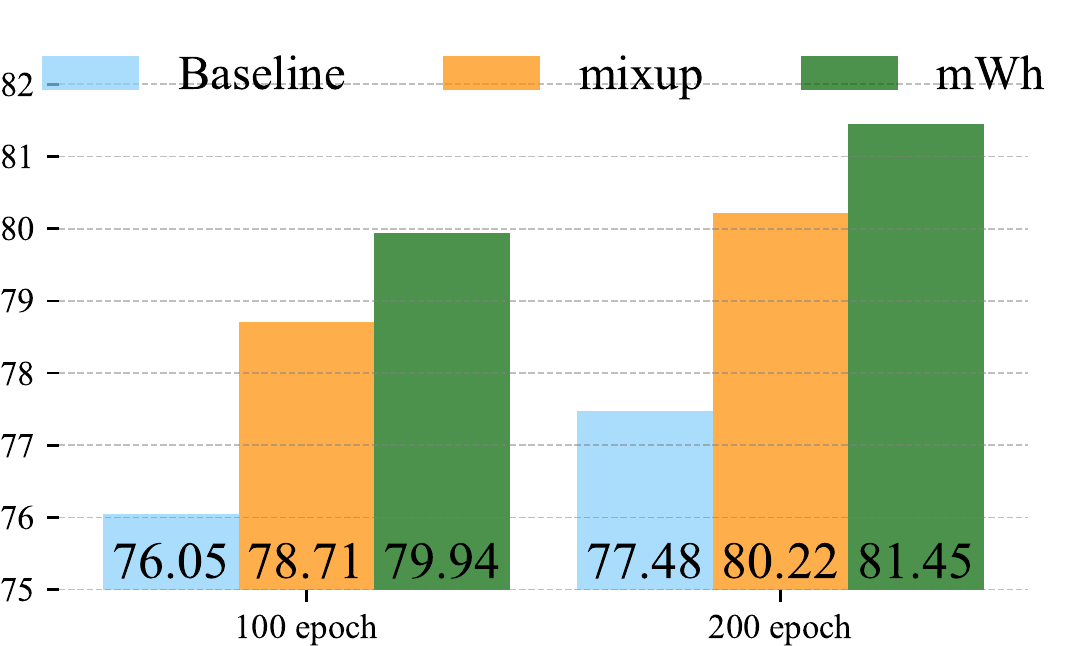}}
  \caption{Accuracy (\%) on CIFAR-10, CIFAR-100 using mixup and mWh.}
\label{Fig1}
  \end{figure*}

From Table \ref{table10}, we can find no matter at what stage, applying mWh instead of mixup leads to higher accuracy. Especially, mixup reduces the performance of the model in Stage 3, which coincides well with results in Table~\ref{table0} and our conjecture (mixup is harmful in later epochs). These results verify the effectiveness of our framework. 

\section{Experiments}
In this section, we evaluate the performance of mWh. We first conduct experiments to validate the effectiveness of mWh on five benchmark classification datasets. For a fair comparison, mWh used the same random seed as mixup. Then we evaluate mWh on COCO2017 detection and show its transferability in CutMix (\citealt{yun2019cutmix}). The parameters $p$ and $q$ are always set to $0.6$ and $0.9$. All our experiments are conducted by PyTorch.

\subsection{Experiments on Image Classification Tasks}
First we will show the results on four small-scale datasets, i.e., CIFAR-10, CIFAR-100 (\citealt{krizhevsky2009learning}), Tiny-ImageNet-200 and CUB-200 (\citealt{welinder2010caltech}). Then the results on ImageNet will be presented. 

We want to show that without doubled epochs and tuning hyper-parameters we can still get better performance in all datasets, so we follow the setting of Zhang et al. (\citeyear{zhang2017mixup}) but halve the training epochs. We also set $\alpha$ to 0.2 and 0.5 to illustrate mWh is robust to hyper-parameter selection. To further show the validity of mWh, we double the epochs and report the results, although it is not our primary focus. 

\textbf{Datasets:}  CIFAR-10 and CIFAR-100 (\citealt{krizhevsky2009learning}) both consist of 50k training and 10k test images at 32x32 resolution. CIFAR-10 contains 5k training and 1k test images per class in a total of 10 classes, and CIFAR-100 has 100 classes containing 600 images each. Tiny-ImageNet-200 contains 200 classes. Each class has 500 training and 50 validation images at 64x64 resolution. For the CUB-200, it contains 200 categories of birds with 5,994 training and 5,794 testing images. The large-scale ImageNet ILSVRC-12 dataset consists of 1.28M training and 50K validation images of various resolutions.

\textbf{Implementation details:} For CIFAR-10 and CIFAR-100, to provide a strong baseline we train DenseNet-121 (\citealt{huang2017densely}) with the mini-batch size of 128 for 100 epochs. Learning rate starts at 0.1 and is divided by 10 after 50 and 75 epochs. Note that we also conduct experiments to double the epoch, i.e., we train the model with 200 epochs, and divide the learning rate by 10 after 100 and 150 epochs.

For Tiny-ImageNet-200, to show that our approach works on a variety of network architectures, in addition to PreAct ResNet-18, we use some lightweight models such as MobileNetV2 (\citealt{sandler2018mobilenetv2}) and EfficientNet-B0 (\citealt{tan2019efficientnet}).  We crop 56x56 patches as input images for training and initialize the learning rate as 0.1. We adopt a mini-batch size 128 and train 200 epochs on Tiny-ImageNet-200.  A step-wise learning rate schedule is applied, that is, the learning rate will be divided by 10 after 75 and 135 epochs. Similarly, the learning rate is decayed by 10-fold at epochs 150, 275 when training 400 epochs.
  
For CUB-200, we use ResNet-18 and crop 224×224 patches as input images for training. For fair comparisons, we evaluate the strategy of training from scratch and fining tune. For training from scratch, we set the number of training epochs to be 175 and 300, and initialize the learning rate as 0.1, batch size as 32. A smoother cosine learning rate adjustment is applied. For fine-tuning, we train the model with 150 and 300 epochs. We set the learning rate as 0.001, batch size as 32. We also use a cosine schedule to scale the learning rate. The initialization ResNet-18 model is downloaded from the PyTorch official website.

To provide further evidence about the quality of representations learned with mWh, we evaluate it on ImageNet. We train ResNet-50 from scratch. For faster convergence we use NVIDIA’s mixed-precision training code base with batch size 2048. The default learning rate is $0.1 *\frac{\text{batch size}}{256}$ with a linear warmup for the first 5 epochs and divided by 10 after training 30, 60, 90 epochs when training 100 epochs, or after 60, 120 and 180 epochs when training 200 epochs. We first randomly crop a patch from the original image and then resize the patch to the target size (224×224). Finally, the patch is horizontally flipped with a probability of 0.5. 

\textbf{Results:} For CIFAR-10 and CIFAR-100, we summarize the results in Figure \ref{Fig1}. mWh consistently outperforms mixup and the baseline. And, mWh with 100 epochs consistently outperforms baseline with even 200 epochs. Note that on CIFAR-100, with a higher $\alpha$, mWh boosts more accuracy. We think the reason is that the difference between the augmented data and the original data will be greater because of a higher $\alpha$, so the empirical risk will decrease more after introducing basic data augmentation. This indicates mWh will bring more improvement with larger $\alpha$, and this situation is particularly noticeable on more complex datasets.
\begin{table}[t]
  \centering
  \small
  \caption{Accuracy (\%) on Tiny-ImageNet-200.}
  \begin{tabular}{c|c|cc|cc}
  \bottomrule [1 pt]
  \multicolumn{2}{c|}{\multirow{2}{*}{Method}}     & \multicolumn{2}{c|}{$\alpha=0.2$} & \multicolumn{2}{c}{$\alpha=0.5$} \\ \cline{3-6} 
  \multicolumn{2}{c|}{}                        & Epochs     & ACC          & Epochs     & ACC          \\ \hline
  \multirow{6}{*}{\begin{tabular}[c]{@{}c@{}}PreAct \\ ResNet-18\end{tabular} }  & Baseline  & 400        & 59.69             & 400        & 59.69             \\
                                    & mixup    & 400        & 60.59             & 400        & 61.58             \\
                                    & mWh     & 400        & \textbf{60.81}    & 400        & \textbf{61.85}    \\ \cline{2-6} 
                                    & Baseline  & 200        & 58.60              & 200        & 58.60              \\
                                    & mixup    & 200        & 60.52             & 200        & 61.44             \\
                                    & mWh     & 200        & \textbf{61.23}    & 200        & \textbf{61.87}    \\ \hline
  \multirow{6}{*}{\begin{tabular}[c]{@{}c@{}}MobileNet \\ V2\end{tabular}}      & Baseline  & 400        & 60.11             & 400        & 60.11             \\
                                    & mixup    & 400        & \textbf{61.83}    & 400        & 61.3              \\
                                    & mWh     & 400        & 61.74             & 400        & \textbf{62.51}    \\ \cline{2-6} 
                                    & Baseline  & 200        & 60.40              & 200        & 60.40              \\
                                    & mixup    & 200        & 60.47             & 200        & 60.59             \\
                                    & mWh     & 200        & \textbf{60.65}    & 200        & \textbf{61.28}    \\ \hline
  \multirow{6}{*}{\begin{tabular}[c]{@{}c@{}}EfficientNet \\ B0\end{tabular}}  & Baseline  & 400        & 52.91             & 400        & 52.91             \\
                                    & mixup    & 400        & 56.17             & 400        & 56.39             \\
                                    & mWh     & 400        & \textbf{56.70}     & 400        & \textbf{57.03}    \\ \cline{2-6} 
                                    & Baseline  & 200        & 53.07             & 200        & 53.07             \\
                                    & mixup    & 200        & 54.49             & 200        & 53.82             \\
                                    & mWh     & 200        & \textbf{55.15}    & 200        & \textbf{54.69}    \\ \toprule[1pt]
  \end{tabular}
  \label{table2}
  \end{table} 
  \begin{table}[!t]
    \caption{Accuracy (\%) on CUB-200. The first two groups of experiments are about training ResNet-18 from scratch and the rest are the fine-tuning experiments. }
    \small
    \centering
    \begin{tabular}{c|cc|cc}
    \bottomrule[1pt]
    \multirow{2}{*}{Method} & \multicolumn{2}{c|}{$\alpha=0.2$} & \multicolumn{2}{c}{$\alpha=0.5$} \\ \cline{2-5} 
                            & Epochs     & Accuracy          & Epochs     & Accuracy          \\ \hline
    Baseline                 & 350    & \multicolumn{1}{l|}{64.308}   & 350          & 64.308          \\
    mixup                   & 350    & \multicolumn{1}{l|}{66.672}   & 350          & 68.347             \\
    mWh                    & 350    & \multicolumn{1}{l|}{\textbf{67.535}}   & 350          & \textbf{70.297}   \\ \hline
    Baseline                 & 175    & \multicolumn{1}{l|}{62.858}   & 175          & 62.858          \\
    mixup                   & 175    & \multicolumn{1}{l|}{\textbf{63.704}}   & 175          & 64.118       \\
    mWh                    & 175    & \multicolumn{1}{l|}{\textbf{63.704}}   & 175          & \textbf{65.948}   \\ \midrule[1pt]
    Baseline                 & 300    & \multicolumn{1}{l|}{77.080}   & 300          & 77.080             \\
    mixup                   & 300    & \multicolumn{1}{l|}{78.650}   & 300          & 78.391        \\
    mWh                    & 300    & \multicolumn{1}{l|}{\textbf{79.272}}   & 300          & \textbf{79.185}   \\ \hline
    Baseline                 & 150    & \multicolumn{1}{l|}{76.345}   & 150          & 76.345    \\
    mixup                   & 150    & \multicolumn{1}{l|}{78.236}   & 150          & 78.219              \\
    mWh                    & 150    & \multicolumn{1}{l|}{\textbf{78.357}}   & 150          & \textbf{78.840}    \\ \toprule[1pt]
    \end{tabular}
    \label{table3}
    \end{table}

\begin{table}[t]
  \centering
  \small
  \caption{Accuracy (\%) on ImageNet with ResNet-50.}
  \begin{tabular}{cl|cc|cc}
  \bottomrule[1pt]
  \multicolumn{2}{c|}{\multirow{2}{*}{Method}} & \multicolumn{2}{c|}{$\alpha=0.2$} & \multicolumn{2}{c}{$\alpha=0.5$} \\ \cline{3-6} 
  \multicolumn{2}{c|}{}                        & Epochs     & Accuracy          & Epochs     & Accuracy          \\ \hline
  \multicolumn{2}{c|}{Baseline}                 & 200        & 76.392              & 200        & 76.392             \\
  \multicolumn{2}{c|}{mixup}                   & 200        & \textbf{77.148}             & 200        & 77.838             \\
  \multicolumn{2}{c|}{mWh}                    & 200        & 77.098    & 200        & \textbf{77.888}    \\ \hline
  \multicolumn{2}{c|}{Baseline}                 & 100        & 76.043              & 100        & 76.043              \\
  \multicolumn{2}{c|}{mixup}                   & 100        & 76.718             & 100        & 76.212             \\
  \multicolumn{2}{c|}{mWh}                    & 100        & \textbf{76.948}     & 100        & \textbf{76.854}      \\ \toprule[1pt]
  \end{tabular}
  \label{table4}
  \end{table}

For Tiny-ImageNet-200, we show the performance comparison in Table \ref{table2}. mWh almost consistently enhances the performance. An exception is when the model is MobileNetV2 and $\alpha$ is 0.2, mWh performs slightly worse than mixup with 400 epochs (61.74 vs. 61.83). In PreAct ResNet-18 and EfficientNet-B0, mWh has a clear advantage than mixup with various $\alpha$. 

Table \ref{table3} shows the results on CUB-200, and similar to the previous experiments, we observe mWh is highly competitive when compared with mixup. When jointly applying mixup and basic data augmentation, mWh obtains the lowest Top-1 error in both training from scratch and fining tune.   

Table \ref{table4} demonstrates  the validation accuracy rates on the ImageNet dataset. mWh outperforms (or is on par with) mixup, mWh also exhibits its robustness to the hyper-parameter $\alpha$. In the 100 epochs case, although mixup is effective when $\alpha = 0.2$, it is not effective when $\alpha = 0.5$ (only improves 0.169\% over the baseline). However, mWh is consistently effective for different $\alpha$ values.
    
As the results have shown, mWh achieves the state-of-the-art performance when halving the epochs, and without deliberately selecting $\alpha$ we can still gain consistently better performance than mixup. Although mWh's goal is not to gain improvement with more epochs, our results show that mWh performs better than mixup when training time is doubled.  These results indicate that without doubling epochs and selecting optimal $\alpha$, mWh can still perform very well. The fact that with more epochs mWh performs better than mixup is a nice byproduct.
\begin{figure*}[t]
  \centering  
  \subfigure[CIFAR-10, $\alpha=0.5$]{
  \label{Fig.sub.1}
  \includegraphics[width=0.23\textwidth]{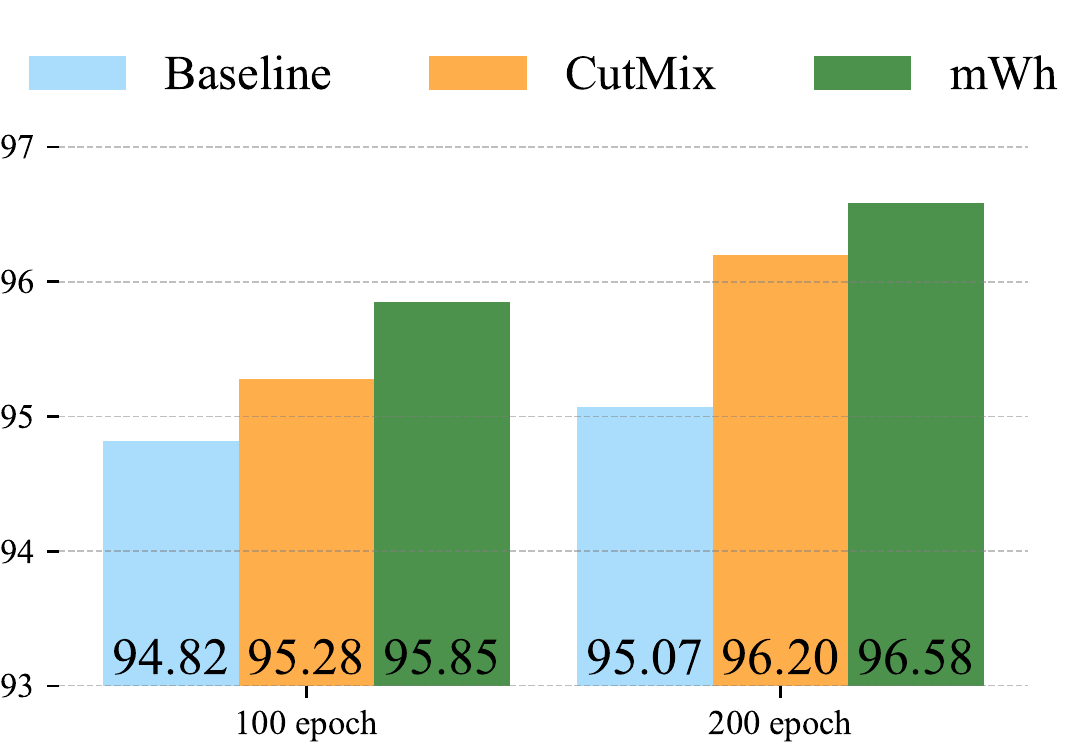}}
  \subfigure[CIFAR-10, $\alpha=1.0$]{
  \label{Fig.sub.2}
  \includegraphics[width=0.23\textwidth]{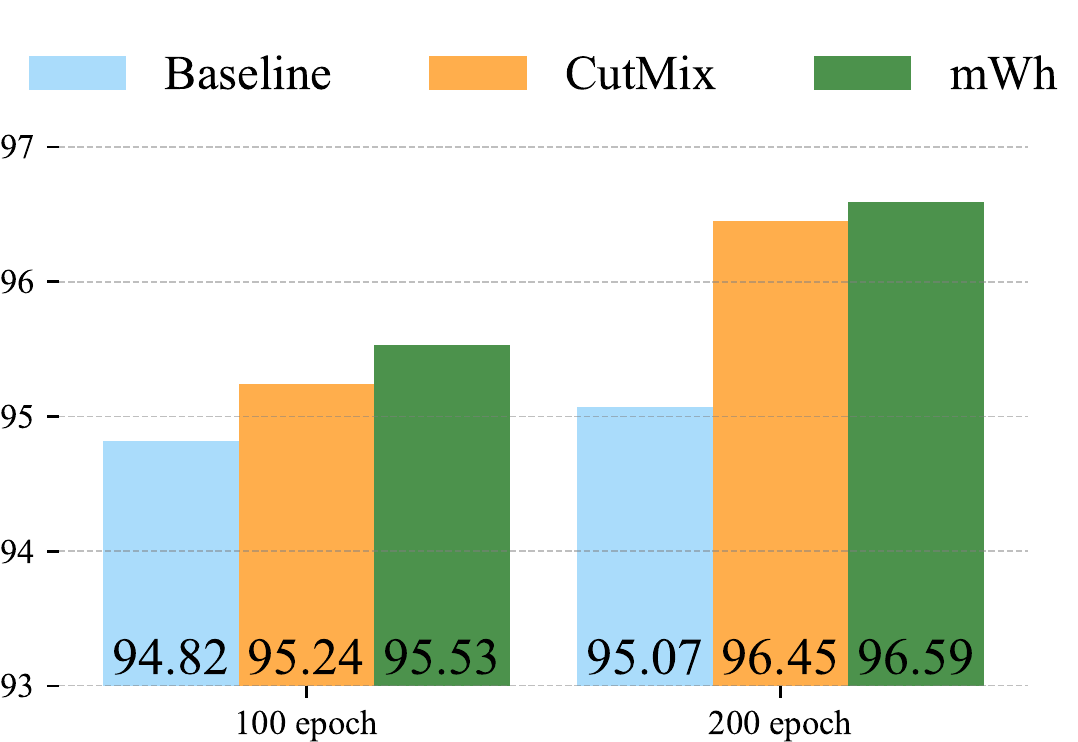}}
  \subfigure[CIFAR-100, $\alpha=0.5$]{
  \label{Fig.sub.1}
  \includegraphics[width=0.23\textwidth]{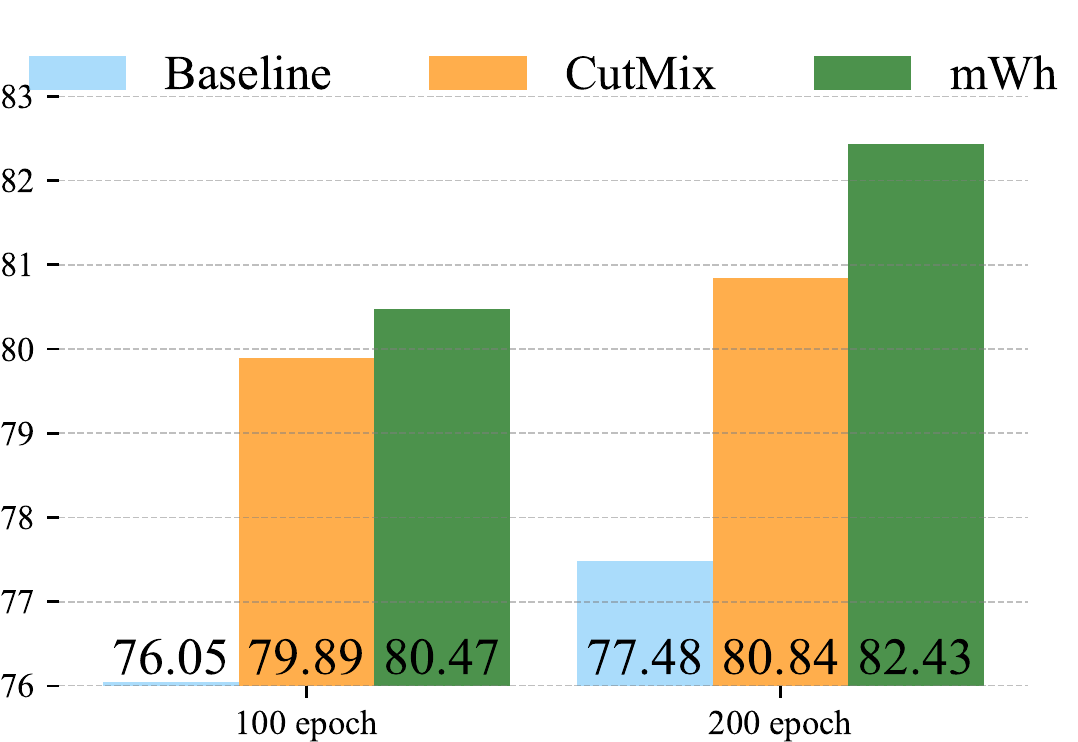}}
  \subfigure[CIFAR-100, $\alpha=1.0$]{
  \label{Fig.sub.2}
  \includegraphics[width=0.23\textwidth]{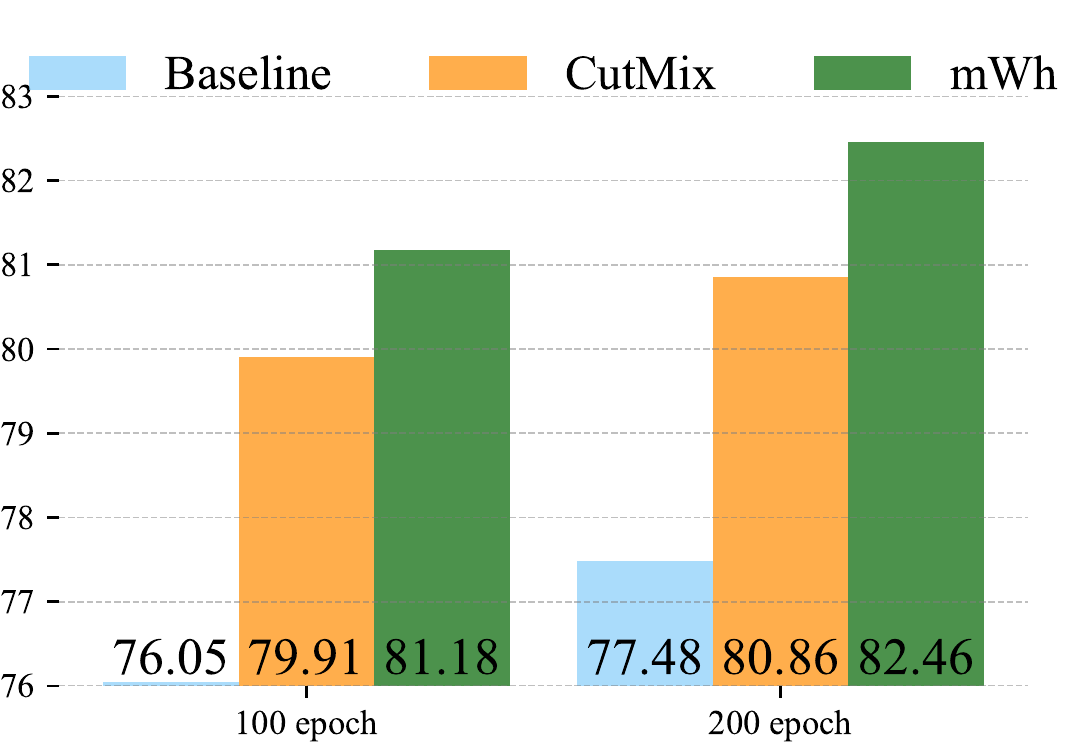}}
  \caption{Accuracy (\%) on CIFAR-10, CIFAR-100 using CutMix and mWh in CutMix.}
\label{Fig3}
  \end{figure*}

\subsection{Transferring to Object Detection}
To validate the effectiveness of mWh on a larger object detection dataset, we conduct experiments on object detection with Faster R-CNN (\citealt{ren2015faster}) on MS COCO2017.

\textbf{Implementation details:} We train Faster R-CNN with ResNet-50 backbone and FPN architecture, we use the PyTorch official torchvision network and the pre-trained ResNet-50 parameters offered by PyTorch official model. We set $\alpha =1.5$. The model is trained for 26 epochs with a mini-batch size of 16, and the learning rate starts at 0.04 and is divided by 10 after 16 and 22 epochs. The data augmentation strategy except for mixup is image horizontally flipping with a probability of 0.5. 

\textbf{Results:} In Table \ref{table5}, the object detection model with mWh succeeds to improve the performance over the original model. mWh without label mixing means that we only mix the image but instead of linearly adding their corresponding labels, and we directly use the label with a larger mixing coefficient to represent the input image.  This strategy also achieves higher mAP than the models with mixup, this shows that the mixing of labels may be unnecessary.
  \begin{table}[t]
    \caption{mAP (\%) on MS COCO 2017 detection, obtained   by learning with the pre-trained ResNet-50 on ILSVRC2012 with Faster
    R-CNN.}
    \centering
  \begin{tabular}{c|c}
  \bottomrule[1pt]
  Methods    & mAP@0.5:0.95  \\   \hline
  Baseline    & 35.5        \\ 
  mixup     & 35.7         \\ 
  mWh & 36.3       \\ 
  mWh without label mixing      & \textbf{36.4}        \\ \toprule[1pt]
  \end{tabular}

  \label{table5}
  \end{table}
\subsection{Classification on Tabular Data }

To further explore the performance of mWh, we evaluate it on tabular data. Zhang et al. (\citeyear{zhang2017mixup}) performed a series of experiments on six classification problems drawn from the UCI repository (\citealt{Dua:2019}), and we follow their settings.

\textbf{Implementation details:} The datasets are scaled from 0 to 1. The neural network is a fully-connected neural network with two hidden layers of 128 ReLU units. The parameters of these neural networks are learned using Adam (\citealt{kingma2014adam}) with default hyper-parameters, and the network is trained for 100 epochs with a mini-batch size of 128.

\textbf{Results:} Table \ref{tablex} shows that mWh improves the average test accuracy on four out of the six considered datasets, and always perform better than mixup. Note that Arrhythmia is a cardiovascular diseases dataset, which has a significant category imbalance problem. This situation may be caused by dividing the dataset inappropriately. Except this datasets, mWh boosts the accuracy in the other five datasets.
\begin{table}[t]
  \centering
  \caption{Accuracy (\%) on the UCI classification datasets.}
  \begin{tabular}{l|lll}
  \bottomrule[1pt]
  Dataset    & Baseline        & mixup  & mWh            \\ \hline
  Abalone    & 25.439         & 25.518 & \textbf{25.997} \\ 
  Arcene     & 81.330          & 81.330  & \textbf{85.330}  \\ 
  Arrhythmia & \textbf{55.735}         & \textbf{55.735} & \textbf{55.735}          \\ 
  Htru2      & 98.026         & 98.038 & \textbf{98.100} \\ 
  Iris       & 93.333 & 95.556 & \textbf{97.780}           \\ 
  Phishing   & 95.799         & 96.281 & \textbf{96.392} \\ \toprule[1pt]
  \end{tabular}
  \label{tablex}
  \end{table}
\subsection{Transferability in CutMix }
In order to provide insights into what makes mWh successful, we further study the transferability of mWh in CutMix. We examine the effect of mWh in CutMix for the image classification tasks. 

\textbf{Implementation details:} All of our following experiments share the same setting with previous ones. We train our strategy in CIFAR, CUB-200 and ImageNet. Note that different from mixup, we select $\alpha$ as 1 instead of 0.2. It is because according to the recommendation of Yun et al. (\citeyear{yun2019cutmix}), choosing $\alpha$ as 1.0 will achieve better performance. For providing strong baselines we set $\alpha$ to 1.0 and 0.5.
\begin{table}[h]
  \caption{ Accuracy (\%) of CutMix and mWh in CutMix on CUB-200. The first two groups of experiments are about training ResNet-18 from scratch and the rest are the fine-tuning experiments. }
  \centering
  \begin{tabular}{c|cc|cc}
  \bottomrule[1pt]
  \multirow{2}{*}{Method} & \multicolumn{2}{c|}{$\alpha=0.5$} & \multicolumn{2}{c}{$\alpha=1.0$} \\ \cline{2-5} 
                          & Epochs     & Accuracy          & Epochs     & Accuracy          \\ \hline
  Baseline                 & 350    & \multicolumn{1}{l|}{64.308}   & 350          & 64.308          \\
  CutMix                   & 350    & \multicolumn{1}{l|}{60.079}   & 350          & 60.563             \\
  mWh                    & 350    & \multicolumn{1}{l|}{\textbf{70.556}}   & 350          & \textbf{70.262}   \\ \hline
  Baseline                 & 175    & \multicolumn{1}{l|}{62.858}   & 175          & 62.858          \\
  CutMix                   & 175    & \multicolumn{1}{l|}{57.007}   & 175          & 56.300       \\
  mWh                  & 175    & \multicolumn{1}{l|}{\textbf{67.691}}   & 175          & \textbf{67.639}   \\ \midrule[1pt]
  Baseline                 & 300    & \multicolumn{1}{l|}{\textbf{77.080}}   & 300          &\textbf{77.080}             \\
  CutMix                   & 300    & \multicolumn{1}{l|}{74.991}   & 300          & 74.456        \\
  mWh                  & 300    & \multicolumn{1}{l|}{75.768}   & 300          & 75.112   \\ \hline
  Baseline                 & 150    & \multicolumn{1}{l|}{76.345}   & 150          & 76.345    \\
  CutMix                   & 150    & \multicolumn{1}{l|}{76.113}   & 150          & 77.045              \\
  mWh                  & 150    & \multicolumn{1}{l|}{\textbf{77.114}}   & 150          & \textbf{78.840}    \\ \toprule[1pt]
  \end{tabular}
  \label{table6}
  \end{table}

\textbf{Results:} The results of CutMix with mWh in CIFAR-10 and CIFAR-100  are shown in Figure \ref{Fig3}. mWh also brings considerable improvement. Table \ref{table6} shows CutMix reduces the accuracy in fine-grained tasks, but mWh in CutMix successfully obtains higher accuracy (or similar ones). In the fine-grained image classification task, only a small area will likely affect the results, and CutMix blurs those features that are important to judge image category. This leaves the model confused by classification, and mWh can eliminate this effect. Table \ref{table8} shows the results of mWh in ImageNet, and our method works well.
\begin{table}[t]
  \caption{Accuracy (\%) on ImageNet comparing CutMix and mWh in CutMix with ResNet-50.  }
  \centering
  \begin{tabular}{cl|cc|cc}
  \bottomrule[1pt]
  \multicolumn{2}{c|}{\multirow{2}{*}{Method}} & \multicolumn{2}{c|}{$\alpha=0.5$} & \multicolumn{2}{c}{$\alpha=1.0$} \\ \cline{3-6} 
  \multicolumn{2}{c|}{}                        & Epochs     & Accuracy          & Epochs     & Accuracy          \\ \hline
  \multicolumn{2}{c|}{Baseline}                 & 100        & 76.043             & 100        & 76.043              \\
  \multicolumn{2}{c|}{CutMix}                   & 100        & 76.591             & 100        & 76.521             \\
  \multicolumn{2}{c|}{mWh}                    & 100        & \textbf{76.734}     & 100        & \textbf{76.774}      \\ \toprule[1pt]
  \end{tabular}
  \label{table8}
  \end{table}
\section{Discussion and Conclusion}

In this paper, we proposed mixup Without hesitation (mWh), a simple but general training policy for effective training. We apply the strategy of reintroducing basic data augmentation to balance exploration and exploitation. Experimental results showed that mWh improves the convergence rate of various dataset instances and is robust to the hyper-parameter selection. It also gains remarkable improvement in different tasks and models compared with the baseline. 

One interesting future work is to investigate the characteristics of mixup that influence the performance of CNN architecture.  In addition, many data augmentation algorithms used in computer vision have similar features as mixup. Therefore, another interesting direction for future research is to extend the proposed algorithm to other augmentation algorithms.

\bibliography{dst_refs}
\bibliographystyle{aaai}
\end{document}